\documentclass[10pt]{article}

\usepackage{fullpage}
\usepackage{setspace}
\usepackage{parskip}
\usepackage{titlesec}
\usepackage[section]{placeins}
\usepackage{xcolor}
\usepackage{breakcites}
\usepackage{lineno}
\usepackage{hyphenat}
\usepackage{mathrsfs}
\usepackage{pdflscape}
\usepackage{rotating}
\usepackage{breakcites}
\usepackage{microtype}

\usepackage[natbibapa]{apacite}

\PassOptionsToPackage{hyphens}{url}
\usepackage[colorlinks = true,
            linkcolor = blue,
            urlcolor  = blue,
            citecolor = blue,
            anchorcolor = blue]{hyperref}
\usepackage{etoolbox}


\usepackage{natbib}

\renewenvironment{abstract}
  {{\bfseries\noindent{\abstractname}\par\nobreak}\footnotesize}
  {\bigskip}

\titlespacing{\section}{0pt}{*3}{*1}
\titlespacing{\subsection}{0pt}{*2}{*0.5}
\titlespacing{\subsubsection}{0pt}{*1.5}{0pt}

\usepackage{authblk}

\usepackage{graphicx}
\usepackage[space]{grffile}
\usepackage{latexsym}
\usepackage{textcomp}
\usepackage{longtable}
\usepackage{tabulary}
\usepackage{booktabs,array,multirow}
\usepackage{amsfonts,amsmath,amssymb}

\newif\iflatexml\latexmlfalse

\AtBeginDocument{\DeclareGraphicsExtensions{.pdf,.PDF,.eps,.EPS,.png,.PNG,.tif,.TIF,.jpg,.JPG,.jpeg,.JPEG}}

\usepackage[utf8]{inputenc}
\usepackage[english]{babel}

\usepackage{float}

\usepackage[margin=1.5in]{geometry}

\usepackage{multirow,bigdelim}
\usepackage{subcaption}
\usepackage{stmaryrd}
\usepackage{bm}

\newcommand{\y}[0]{$\mathbf{\times}$}
\newcommand{\p}[0]{p }
\newcommand{\di}[0]{d }
\newcommand{\med}[0]{med }
\newcommand{\proc}[0]{proc }
\newcommand{\pres}[0]{med }
\newcommand{\clinical}[0]{cl }
\newcommand{\lab}[0]{lab }
\newcommand{\sg}[0]{sg }
\newcommand{\diverse}[0]{cl }

\newcommand*\row[9]{%
  \begingroup
  \def\tempa{#1}%
  \def\tempb{#2}%
  \def\tempc{#3}%
  \def\tempd{#4}%
  \def\tempe{#5}%
  \def\tempf{#6}%
  \def\tempg{#7}%
  \def\temph{#8}%
  \def\tempi{#9}%
  \foocont
}
\newcommand*\foocont[2]{%
  \edef\processme{%
    \endgroup
    \unexpanded\expandafter{\tempa} &
    \unexpanded\expandafter{\tempb} &
    \unexpanded\expandafter{\tempc} &
    \unexpanded\expandafter{\tempd} &
    \unexpanded\expandafter{\tempe} &
    \unexpanded\expandafter{\tempf} &
    \unexpanded\expandafter{\tempg} &
    \unexpanded\expandafter{\temph} &
    \unexpanded\expandafter{\tempi} & #1 & #2\cr
  }%
  \processme
}

\newcommand{\Tra}{^{{\sf T}}} 
\newcommand{\V}[1]{{\bm{\mathbf{\MakeLowercase{#1}}}}} 

\newcommand{\M}[1]{{\bm{\mathbf{\MakeUppercase{#1}}}}} 
\newcommand{\Mhat}[1]{{\bm{\hat \mathbf{\MakeUppercase{#1}}}}} 
\newcommand{\Mbar}[1]{{\bm{\bar \mathbf{\MakeUppercase{#1}}}}} 
\newcommand{\ME}[2]{\MakeLowercase{#1}_{#2}} 
\newcommand{\MC}[2]{\V{#1}_{#2}} 

\newcommand{\T}[1]{\mathcal{#1}} 

\newcommand{\KOp}[1]{\llbracket #1 \rrbracket} 

\newcommand{\norm}[1]{\left\lVert \, #1 \, \right\rVert}

\newcommand{\Sec}[1]{\hyperref[sec:#1]{\S\ref*{sec:#1}}} 
\newcommand{\Section}[1]{\hyperref[sec:#1]{Section~\ref*{sec:#1}}} 
\newcommand{\AppFull}[1]{\hyperref[sec:#1]{Appendix~\ref*{sec:#1}}} 
\newcommand{\Eqn}[1]{\hyperref[eq:#1]{(\ref*{eq:#1})}} 
\newcommand{\Fig}[1]{\hyperref[fig:#1]{Figure~\ref*{fig:#1}}} 
\newcommand{\Tab}[1]{\hyperref[tab:#1]{Table~\ref*{tab:#1}}} 
\newcommand{\Thm}[1]{\hyperref[thm:#1]{Theorem~\ref*{thm:#1}}} 
\newcommand{\Cor}[1]{\hyperref[cor:#1]{Corollary~\ref*{cor:#1}}} 
\newcommand{\Alg}[1]{\hyperref[alg:#1]{Algorithm~\ref*{alg:#1}}} 
\newcommand{\Def}[1]{\hyperref[def:#1]{Definition~\ref*{def:#1}}} 

\newcommand{\Real}{{\mathbb R}}

\iflatexml

\else
\fi

\begin{document}
\bibliographystyle{apacite}

\title{Unsupervised EHR-based Phenotyping via Matrix and Tensor Decompositions}

\author[1,2]{Florian Becker}%
\author[1,3]{Age K. Smilde}%
\author[1]{Evrim Acar}%

\affil[1]{Simula Metropolitan Center for Digital Engineering, Oslo, Norway}%
\affil[2]{Oslo Metropolitan University, Oslo, Norway}
\affil[3]{Swammerdam Institute for Life Sciences, University of Amsterdam, Amsterdam, Netherlands}

\vspace{-1em}

  \date{}

\begingroup
\let\center\flushleft
\let\endcenter\endflushleft
\maketitle
\endgroup

\selectlanguage{english}

\begin{abstract}
Computational phenotyping allows for unsupervised discovery of subgroups of patients as well as corresponding co-occurring medical conditions from electronic health records (EHR). Typically, EHR data contains demographic information, diagnoses and laboratory results. Discovering (novel) phenotypes has the potential to be of prognostic and therapeutic value. Providing medical practitioners with transparent and interpretable results is an important requirement and an essential part for advancing precision medicine. Low-rank data approximation methods such as matrix (e.g., non-negative matrix factorization) and tensor decompositions (e.g., CANDECOMP/PARAFAC) have demonstrated that they can provide such transparent and interpretable insights. Recent developments have adapted low-rank data approximation methods by incorporating different constraints and regularizations that facilitate interpretability further. In addition, they offer solutions for common challenges within EHR data such as high dimensionality, data sparsity and incompleteness. Especially extracting \emph{temporal phenotypes} from longitudinal EHR has received much attention in recent years. In this paper, we provide a comprehensive review of low-rank approximation-based approaches for computational phenotyping. The existing literature is categorized into temporal vs. static phenotyping approaches based on matrix vs. tensor decompositions. Furthermore, we outline different approaches for the validation of phenotypes, i.e., the assessment of clinical significance. 
\end{abstract}


\maketitle

\section{Introduction}

The fast adoption of healthcare information systems and the resulting aggregation of electronic health records (EHR) has led to the development of different $\textit{phenotyping}$ methods with the goal of identifying co-occurring medical conditions and discovering subgroups of patients that, for instance, share certain disease (progression) characteristics. EHR are typically comprised of demographic data, diagnoses, laboratory test results and prescriptions. The discovery of patient subgroups was formerly based on medical expertise or by heuristics. Computational phenotyping accelerates and facilitates this process. As labeling EHR data is very labor-intensive \citep{hripcsak2013next}, and one of the main goals of computational phenotyping is to identify novel phenotypes, effective unsupervised methods are needed. These automated and unsupervised approaches are highly relevant for medical practitioners as well as researchers because of the potential of transforming idiosyncratic, raw and unlabeled health records into clinically relevant, explainable and interpretable concepts \citep{ho2014marble, shivade2014review}. Thus, the discovery and assessment of phenotypes might improve the understanding of a disease and can be of prognostic or therapeutic value thereby promoting personalized medicine \citep{abul2019personalized}. 

\sloppy{The trend from traditional rule-based phenotyping systems that require medical expertise towards data-driven methods is ongoing \citep{shivade2014review, richesson2016clinical}. Natural language processing (NLP) has been playing a crucial role to extract phenotypes and clinical relationships or features from unstructured EHR such as clinical notes \citep{banda2018advances, friedman1999natural}. Phenotypes have been extracted from EHR using unsupervised methods such as various clustering approaches \citep{doshi2014comorbidity, pikoula2019identifying, ScWi15}. 
Deep learning (DL) methods have become increasingly popular for the (mainly supervised) analysis of both static and temporal EHR data \citep{solares2020deep}. However, the lack of interpretability and \textit{model transparency} has been repeatedly raised as the main disadvantage of DL-based methods \citep{shickel2017deep, xiao2018opportunities}.}

\begin{figure*}[t!]
\centering
\includegraphics[width=0.9\textwidth]{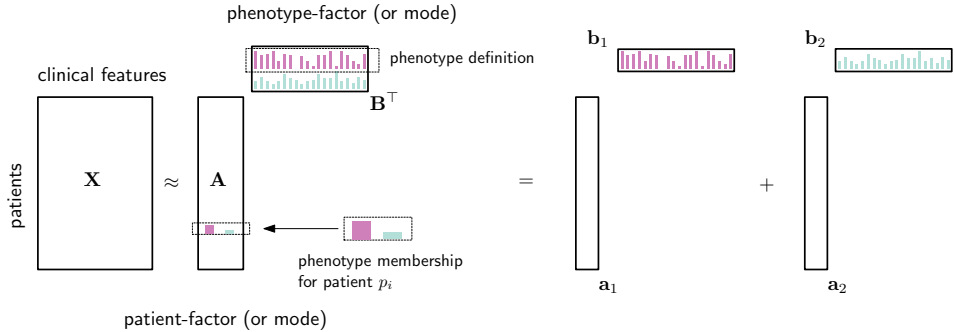}
\caption{A \emph{patients} $\times$ \emph{clinical features} (e.g., laboratory measurements) matrix is approximated using patient-mode factors and phenotype-mode factors (that reveal phenotypes). What is shown here is a schematic factorization using a rank-2 model.}\label{fig:matrix-decomp}
\end{figure*}

On the other hand, low-rank data approximations (matrix factorizations as well as tensor factorizations, i.e., extensions of matrix factorizations to higher-order data sets) have shown promise in terms of providing medical practitioners and researchers with interpretable results \citep{luo2017tensor,banda2018advances}. Low-rank structures typically arise due to similar patient groups sharing a set of frequently co-occurring medical conditions. For instance, EHR data can be arranged as a \emph{patients} by \emph{clinical features} matrix (as in Figure \ref{fig:matrix-decomp}), and a low-rank approximation of the matrix summarizes the data using only several factors (much smaller than the number of features) corresponding to various phenotypes. Since EHR data is quite rich and often multi-way, i.e., has more than two modes of variation, the data can be represented as a multi-way array (also referred to as a higher-order tensor), e.g., \emph{patients} by \emph{diagnoses} by \emph{medications} tensor as in Figure \ref{fig:static-tensor} or \emph{patients} by \emph{clinical features} by \emph{time} tensors as in Figure \ref{fig:dynamic-tensor}. Similar to matrix factorizations, tensor factorizations can provide interpretable summaries of such multi-way data revealing phenotypes. Matrix and tensor factorizations also allow for a variety of constraints, for instance, to enforce more distinct or sparse phenotypes; thereby, facilitating clinical analysis and interpretation.

As a result of their effectiveness in terms of revealing interpretable patterns from complex EHR data in an unsupervised way, matrix and tensor factorizations have been widely studied in recent years. Especially temporal phenotyping via tensor decompositions has received much attention \citep{perros2017spartan, afshar2018copa, perros2019temporal,zhao2019detecting, zhang2021feature}. Temporal phenotyping focuses on the discovery of phenotypes as well as their temporal changes through the analysis of longitudinal EHR data. Extracting \textit{temporal signatures} of phenotypes can contribute to the general understanding of a disease and provide more accurate phenotypes \citep{perros2017spartan, zhang2021feature}. Here, we review matrix and tensor factorization-based approaches for electronic health records-based phenotyping, discuss their strengths and limitations, as well as possible future research directions.

While there are recent survey papers on computational phenotyping reviewing NLP methods \citep{zeng2018natural}, deep learning-based approaches \citep{shickel2017deep, solares2020deep}, giving a more general overview over different data mining and machine learning techniques \citep{shivade2014review}, and discussing the general transition from rule-based systems towards machine learning models \citep{banda2018advances}, to the best of our knowledge, tensor methods have only been briefly discussed for precision medicine \citep{luo2017tensor}.
\sloppy{In this paper, we provide a comprehensive review of low-rank approximation-based approaches for unsupervised phenotyping categorizing the existing literature as temporal vs.~static phenotyping approaches based on matrix vs. tensor decompositions. Moreover, we review validation approaches crucial for unsupervised phenotyping. Within the context of this survey, we discuss the key challenges in phenotype discovery - identified and addressed in the literature: high dimensionality \citep{perros2017spartan, perros2019temporal, yang2017tagited}, data sparsity \citep{zhou2014micro, perros2017spartan, schuler2016discovering}, incompleteness \citep{zhou2014micro, yin2020logpar, perros2019temporal, schuler2016discovering, ren2020robust}, irregularity along the time mode \citep{ren2020robust,yin2020logpar, afshar2018copa, yin2019learning,  perros2019temporal}, overlapping phenotypes \citep{henderson2017granite, ho2014marble, kim2017discriminative, yin2020learning}, different statistical data types \citep{yang2017tagited, perros2018sustain, schuler2016discovering} and privacy-preserving computational phenotyping \citep{ma2019privacy, kim2017federated}.}

Using a hierarchical categorization, we first distinguish between matrix vs.~tensor factorizations-based approaches. Then, both categories are further subdivided into static vs.~temporal phenotyping. Section  \ref{sec:matrixfactorizations} and \ref{sec:tensorfactorizations} will follow this structure after a brief introduction of commonly used matrix and tensor factorizations in EHR-based phenotyping in Section \ref{sec:background}. Table \ref{tab:literature}  shows different aspects under which a more fine-grained categorization of the existing literature is possible. For example, the approaches to validate phenotypes differ greatly (Section \ref{sec:validation}). This survey covers papers that have their emphasis on methodological development for phenotyping (e.g., \citep{ho2014marble,yin2020learning,yin2020logpar}) as well as papers that focus on applications using well-established low-rank approximation-based methods (e.g., \citep{aliberti2016clinical}). 

\section{Background: Phenotyping via Low-Rank Approximations}\label{sec:background}

\subsection{Data: From EHR to Matrices and Tensors}

Extracting and categorizing information from unstructured EHR (e.g., clinical notes) is a research topic in itself. Both classical NLP models \citep{khalifa2015adapting} and DL-based methods \citep{shickel2017deep} have been used to transform raw EHR to concepts and relations, such as prescriptions, and diagnoses. 
Matrices and higher-order tensors can be naturally used to represent co-occurrences of patients, procedures, diagnoses, and/or prescriptions (over time). 
For instance, EHR data can be arranged as a \emph{patients} by \emph{clinical features} matrix as in Figure \ref{fig:matrix-decomp} \citep{rennard2015identification, georgiades2007structure}, or given that different diagnoses and medications are the available data points for a cohort of patients, the data can be arranged as a \emph{patients} by \emph{diagnoses} by \emph{medications} tensor (Figure \ref{fig:static-tensor}) \citep{henderson2018phenotyping, ho2014limestone, kim2017discriminative}. Consider, for instance, a patient $i$ that is prescribed a medication $k$ for a certain condition $j$. This information can be encoded in a 3-way array by setting the corresponding entry $(i,j,k) = 1$.
In this way, by linking medications and conditions, \textit{multi-modal} interactions can be encoded. Tensors also prove useful in terms of representing temporal data, e.g., multiple visits of patients can be recorded using an irregular tensor with modes: \emph{patients}, \emph{clinical features}, and \emph{time} (Figure \ref{fig:dynamic-tensor}) \citep{perros2017spartan, yin2020logpar, ren2020robust}. Table \ref{tab:literature} shows matrix and higher-order tensor representations of various types of EHR data in the literature. 

For an overview of commonly used data sets in the literature, see Table \ref{tab:datasets}. The MIMIC-III (Medical Information Mart for Intensive Care) dataset, for instance, that contains health records of intensive care unit (ICU) patients, is used in several  phenotyping studies based on matrix/tensor factorizations (e.g., \citep{joshi2016identifiable, yin2020learning, ma2019privacy, yin2019learning}). Diagnosis codes are typically given according to the International Classification of Diseases (ICD). Resulting co-occurrences in the form of, e.g., \textit{patient-diagnosis-prescription}, are encoded using either binary values \citep{yin2020logpar} or counts \citep{ho2014marble, ho2014limestone, luo2016predicting, ding2021unsupervised}. Temporal scales (time between observations) vary a lot in different studies, i.e., from weeks \citep{perros2017spartan, zhou2014micro} to months \citep{schuler2016discovering} and years \citep{zhao2019detecting}.

\subsection{Preliminaries of Low-Rank Approximations}

Low-rank approximations are effective tools for revealing the underlying patterns from data in the form of matrices and higher-order tensors. 
In the following, we briefly review the basics of low-rank approximations using matrix and tensor factorizations, and link them to (temporal) EHR-based phenotyping.

\subsubsection{Matrix Decompositions}
Given a data matrix $\mathbf{X} \in \mathbb{R}^{I \times J}$ in the form of a \emph{samples} by \emph{features} matrix, it can be approximated using a matrix factorization as follows:
\begin{equation}
\label{eqn:MF}
\centering
\mathbf X \approx \M{A}\M{B}\Tra ,
\end{equation}
where $\M{A} = [\MC{A}{1} \hspace{0.05in} ... \hspace{0.05in} \MC{A}{R}] \in \Real^{I \times R}$ and $\M{B}= [\MC{B}{1} \hspace{0.05in} ... \hspace{0.05in} \MC{B}{R}]\in \Real^{J \times R}$ are factor matrices corresponding to the \emph{samples} and \emph{features} modes summarizing the data matrix using $R$ components, with $R \ll \min\{I,J\}$. Alternatively, the formulation (\ref{eqn:MF}) can be expressed as the sum of $R$ rank-one terms/components, i.e.,
$\mathbf X \approx \sum_{r=1}^{R} \mathbf a_r \mathbf b_r\Tra$. 
Once the summarization of the data is obtained using such a low-rank approximation, factor matrices can be used for further analysis, e.g., if $\M{X}$ is a \emph{patients} by \emph{clinical features} matrix (as in Figure \ref{fig:matrix-decomp}), the factor matrix $\M{A}$ corresponding to the \emph{patients} mode can be used to cluster the patients. In the literature, $\M{A}$ is also referred to as the \textit{score matrix}, and $\M{B}$ as the \textit{loading matrix} \citep{BrSm14}, and in phenotyping literature, $\M{A}$ is sometimes also called the membership matrix \citep{ho2014limestone, henderson2017granite, henderson2018phenotyping} or patient mode, or patient factor matrix \citep{ma2019privacy}. We use these terms interchangeably.

Given a matrix $\M{X}$, factor matrices $\M{A}$ and $\M{B}$ can be computed by solving the following optimization problem:
\begin{equation}
   \min_{\mathbf A, \mathbf B } \norm{\smash{\M{X} - \M{A}\M{B}\Tra}}_F^2,
\end{equation}
where $\norm{.}_F$ denotes the Frobenius norm, i.e., $\norm{\M{X}}_F = \sqrt{\sum_{i=1}^I \sum_{j=1}^J \ME{X}{ij}^2}$.

Imposing different constraints on the factor matrices $\M{A}, \M{B}$ leads to different matrix factorization approaches. For instance, in Singular Value Decomposition (SVD), which can be used for Principal Component Analysis (PCA) \citep{JoCa16}, factor matrices are constrained to have orthogonal columns. By imposing non-negativity constraints on the factor matrices, the problem can be formulated as a non-negative matrix factorization (NMF) \citep{lee1999learning} problem. 
Note that the factorization of $\M{X}$ as $\M{X} \approx \M{A}\M{B}\Tra$ is not unique as factor matrices can be multiplied by a nonsingular matrix $\M{M}$ and its inverse, such that $\M{A}\M{B}\Tra= \M{A}\M{M}\M{M}^{-1}\M{B}\Tra= \Mbar{A}\Mbar{B}\Tra$ giving an equally good approximation of $\M{X}$, where  $\Mbar{A} = \M{A}\M{M}$ and $\Mbar{B}=\M{B}(\M{M}^{-1})\Tra$. If the goal of data approximation using matrix factorization is to reveal patterns, i.e., columns of factor matrices, and interpret the individual patterns, for instance, as phenotypes, then the factorization must be unique. Therefore, often additional constraints on the factors such as orthogonality, non-negativity and sparsity or statistical independence are needed to obtain unique factorizations. Here, with uniqueness, we refer to essential uniqueness which means that the factorization is unique up to permutation of rank-one components, and scaling within each rank-one component.

The Frobenius norm-based loss function is common in data mining, and relies on the assumption that data is real-valued. On the other hand, for EHR data analysis and phenotyping applications, other types of loss functions may be more suitable, e.g., when analyzing count data, a better suited loss function is based on \emph{Kullback-Leibler (KL) divergence} relying on the Poisson distribution. The matrix factorization problem with different loss functions can be formulated as follows:
\begin{align}
\begin{split}
\label{eq:KL}
   \min_{\M{A},\M{B}} & \quad d(\M{X},\Mhat{X})\\
   \textrm{s.t.} & \quad \Mhat{X}=\M{A}\M{B}\Tra
\end{split}
\end{align}
where $d(\M{X},\Mhat{X})$ indicates the loss function, e.g., for KL-divergence, $d(\M{X},\Mhat{X})= \sum_{i}^I \sum_{j}^J d(x_{ij}, \hat{x}_{ij})$, with $d(x, y) = x \textrm{log}\frac{x}{y} - (x-y)$, for $x,y > 0$. Recently, \emph{generalized low-rank models} (GLRM) \citep{udell2016generalized} have been introduced to model data matrices, where different columns may have different statistical data types (e.g., binary, count, real), and each column can be modelled using a suitable loss function.


Table \ref{tab:literature} shows representative loss functions commonly used with matrix factorizations in the phenotyping literature. 


\subsubsection{Tensor Decompositions}

Higher-order \textit{tensors} (or multi-way arrays) are  higher-order extensions of data matrices. We use $\mathcal{X} \in \mathbb{R}^{I_1, \dots, I_K}$ to denote a tensor of \textit{order} $K$. While vectors and matrices are special cases of tensors for $K=1$ and $K=2$, respectively, we use the term \textit{tensor} to refer to the case $K>2$. Tensor factorizations are extensions of matrix factorizations to such higher-order data sets, and have been successfully used to extract underlying patterns in multi-way data in many disciplines including neuroscience \citep{AcBiBiBr07, HuDuPa17, WiKiWa18}, social network analysis \citep{AcDuKo10b, PaFaSi16} and chemometrics \citep{bro1997parafac}. Among various tensor factorization approaches, here, we briefly discuss the CANDECOMP/PARAFAC (CP) \citep{harshman1970foundations, carroll1970analysis} and PARAFAC2 \citep{harshman1972parafac2, kiers1999parafac2} models as they have been widely used in the phenotyping literature. Uniqueness properties of CP and PARAFAC2, i.e., both being unique up to permutation and scaling ambiguities under certain conditions \citep{kruskal1977three,kiers1999parafac2}, make them powerful tools in terms of discovering phenotypes compared to other tensor methods. See  \citep{smilde2005multi, acar2008unsupervised, kolda2009tensor, PaFaSi16} for thorough reviews on tensor methods and their applications in different fields.

\paragraph{CANDECOMP/PARAFAC (CP)} 
The CP model \citep{harshman1970foundations,carroll1970analysis}, also referred to as the canonical polyadic decomposition \citep{hitchcock1927expression}, represents tensor $\mathcal{X}$ as the sum of minimum number of rank-one components (see Figure \ref{fig:static-tensor}). Given a tensor $\mathcal{X} \in \Real^{I \times J \times K}$, its $R$-component CP model can be expressed as follows: 
\begin{equation}
   \mathcal X \approx \sum_{r=1}^{R} \mathbf a_r \, \circ \, \mathbf b_r \, \circ \, \mathbf c_r,
\end{equation}
where $\circ$ denotes the vector outer product, $\M{A} = [\MC{A}{1} \hspace{0.05in} ... \hspace{0.05in} \MC{A}{R}] \in \Real^{I \times R}$, $\M{B}= [\MC{B}{1} \hspace{0.05in} ... \hspace{0.05in} \MC{B}{R}]\in \Real^{J \times R}$, and $\M{C}= [\MC{c}{1} \hspace{0.05in} ... \hspace{0.05in} \MC{c}{R}]\in \Real^{K \times R}$ are the factor matrices corresponding to each mode. The CP model is often denoted as $\mathcal X \approx  \KOp{\M{A},\M{B},\M{C}}$. If $\mathcal{X}$ is a \emph{patients} by \emph{diagnoses} by \emph{medications} tensor, factor matrices $\M{A}, \M{B}, \M{C}$ correspond to the patients, diagnoses and medications modes (see Figure \ref{fig:static-tensor}). Here,  $\M{A}$ can be used as the phenotype membership matrix, and columns of $\M{B}$ and $\M{C}$ can be used to reveal phenotypes.

The CP model has been an effective tool in computational phenotyping as a result of its uniqueness properties, which enables interpreting columns of factor matrices as phenotypes or patient subgroups. The CP model is essentially unique, i.e., factor matrices are unique up to scaling and permutation ambiguities, under mild conditions \citep{kruskal1977three, sidiropoulos2000uniqueness} without the need to impose any additional constraints such as orthogonality or statistical independence on the columns of factor matrices. The scaling ambiguity indicates that individual vectors in each rank-one component can be scaled as in: 
\begin{equation*}
   \mathcal X \approx \sum_{r=1}^{R} \mathbf a_r \, \circ \, \mathbf b_r \, \circ \, \mathbf c_r = \sum_{r=1}^{R}  \alpha \, \mathbf a_r \, \circ \,  \beta \, \mathbf b_r \, \circ \,  \gamma \, \mathbf c_r,
\end{equation*}
as long as $\alpha\beta\gamma=1$. The permutation ambiguity refers to different ordering of rank-one components. Neither scaling nor permutation ambiguity changes the interpretation of the factors.

An alternative way to represent the CP model, in matrix notation, is as follows:
\begin{equation}
   \M{X}_k \approx \mathbf A \operatorname{diag}(\mathbf c_{k:}) \mathbf B^{\intercal},
\end{equation}
where $\M{X}_k$ corresponds to $k$th frontal slice of tensor $\mathcal X$, and $\operatorname{diag}(\mathbf c_{k:})$ denotes an $R \times R$ diagonal matrix with the $k$th row of $\M{C}$ as diagonal entries. 
Given a tensor $\mathcal X$, the CP model can be computed by solving the following optimization problem using alternating least squares (ALS) \citep{harshman1970foundations} or all-at-once optimization based approaches \citep{AcDuKo11a,SoBaLa13a}:
\begin{equation} \label{eq:CP}
   \min_{\M{A}, \M{B}, \M{C}} \norm{\mathcal{X} -\KOp{\M{A},\M{B},\M{C}}}_{F}^2.
\end{equation}

Similar to matrix factorizations, loss functions other than the squared Frobenius norm, in particular, KL-divergence for count data, have been commonly used in the phenotyping literature (see Table \ref{tab:literature}). We mention here the corresponding optimization problem due to its prevalence in static phenotyping.
Note that the following KL-formulation differs from the one given in (\ref{eq:KL}) as it omits the constants (i.e., terms that only include the data $x_{ijk}$ and have no impact on the optimization). Let $\mathcal{\hat{X}}$ be the approximation of $\mathcal X$ by the factor matrices based on a CP model, and let $x_{ijk}$ be the corresponding entry in tensor $\mathcal X$, then:
\begin{align*}
   \min_{\mathbf A, \mathbf B, \mathbf C} \quad & \sum_{i=1}^I \sum_{j=1}^J \sum_{k=1}^K  \hat{x}_{ijk} - x_{ijk} \log \hat{x}_{ijk} \\
\textrm{s.t.} \quad & \hat{\mathcal{X}} = \KOp{\mathbf A, \mathbf B, \mathbf C } \\ 
\end{align*}

\paragraph{PARAFAC2} While the CP model assumes that each frontal slice (or \textit{slab}) is approximated by the same $\mathbf{B}$ matrix, the PARAFAC2 model \citep{harshman1972parafac2, kiers1999parafac2} relaxes this assumption, and allows the $\mathbf{B}$ matrix to change across different slabs as follows: 
\begin{equation}
\label{eq:Parafac2}
   \mathbf X_k \approx \mathbf A \operatorname{diag}(\mathbf c_{k:}) \mathbf B_k^{\intercal},
\end{equation}
subject to the so-called \textit{PARAFAC2 constraint}, i.e.,~ 
\begin{equation}
 \mathbf B_{k_1}\Tra \mathbf B_{k_1} = \mathbf B_{k_2}\Tra\mathbf B_{k_2} = \mathbf{\Phi}, \quad \forall \, k_1, k_2 \leq K.
\end{equation}

This constraint, which enforces invariance over the cross products, was introduced to preserve uniqueness \citep{harshman1972parafac2}. Similar to CP, the factors matrices from a PARAFAC2 decomposition are essentially unique under mild conditions, e.g., as long as there are enough slices ($K > 3$) \citep{kiers1999parafac2}. In addition to permutation and scaling ambiguities, we may encounter an additional sign ambiguity when fitting a PARAFAC2 model, where each entry in the $k$th row of $\M{C}$ in Eqn. \ref{eq:Parafac2} may flip signs arbitrarily. One possible solution to handle the sign ambiguity is to impose non-negativity constraints on matrix $\M{C}$ \citep{harshman1972parafac2, kiers1999parafac2}. 

\begin{figure*}[ht]
\centering
\includegraphics[width=0.6\textwidth]{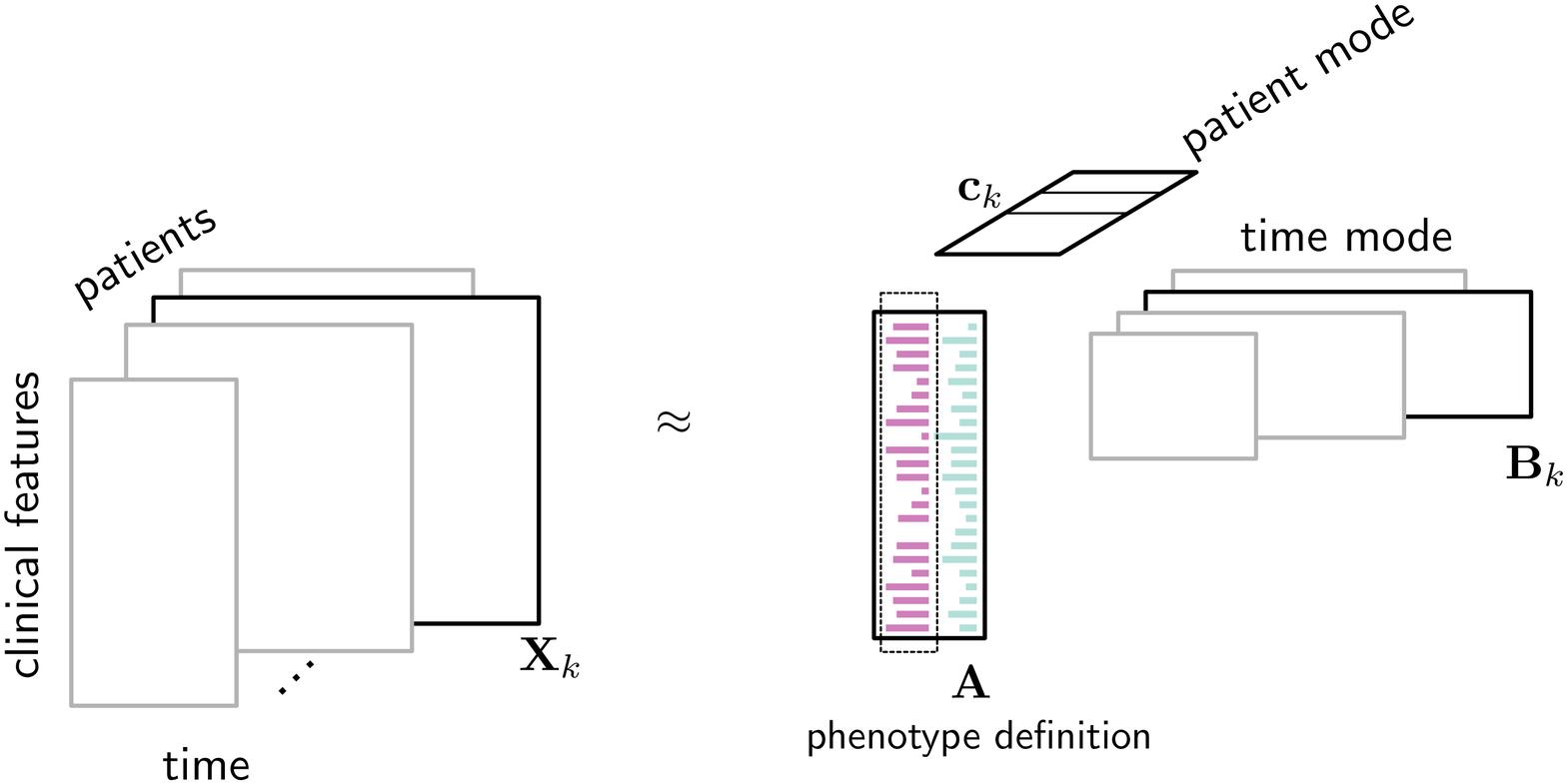}
\caption{Temporal phenotyping. An irregular \emph{patients} $\times$ \emph{clinical features} $\times$ \emph{time} tensor is decomposed into the corresponding three modes: patients, phenotype, time.  Irregularity in the time mode arises, for example, due to a varying number clinical visits of the cohort. The PARAFAC2 model is able to factorize irregular tensors.}
\label{fig:dynamic-tensor}
\end{figure*}

Allowing each slab to have its corresponding $\mathbf B_k$ also means that irregular $\mathbf X_k \in \mathbb{R}^{I \times J_k}$ can the modeled, i.e., the dimension ($J_k$) does not necessarily have to align across slabs. This property of PARAFAC2 is particularly useful for the analysis of EHR data as it is often the case that clinical visits (or the time mode in general) do not align across different patients. For instance, if $\M{X}_k$, for $k=1, ..., K$, corresponds to a \emph{clinical features} by \emph{time/visit} matrix for patient $k$, the factor matrix $\M{A}$ reveals phenotypes together with their corresponding temporal profiles as the columns of factor matrix $\M{B}_k$ for patient $k$, while the factor matrix $\M{C}$ corresponds to the patient mode potentially revealing patient groups (see Figure \ref{fig:dynamic-tensor}).

The PARAFAC2 model can be computed by solving the  following optimization problem using an ALS-based algorithm \citep{kiers1999parafac2}:
\begin{equation*}
   \min_{\M{A},  \{\M{B}_k\}_{k \leq K}, \M{C}}  \sum_{k=1}^K \norm{\M{X}_k -\M{A} \operatorname{diag}(\mathbf c_{k:}) \M{B}_k^{\intercal}}_{F}^2,
\end{equation*}
where $\M{B}_k = \M{P}_k\M{B}$, and $\M{P}_k\Tra\M{P}_k=\M{I}$ so that the constant cross product constraint is implicitly satisfied. Here, $\M{I} \in \Real^{R \times R}$ denotes the identity matrix, $\M{B} \in \Real^{R \times R}$ is common for all $\M{B}_k$, $k=1, ..., K$.
Other algorithmic approaches have also been studied when constraints are needed on the factor matrices \citep{CoBr18,RoScBr22}. Within the context of temporal phenotyping, PARAFAC2 has been extended to model binary data \citep{yin2020logpar}. However, the standard formulation based on the Frobenius norm has so far been the most commonly used (see Table \ref{tab:literature}).

\section{Matrix Factorizations for Phenotyping}
\label{sec:matrixfactorizations}

In this section, matrix factorization-based approaches are categorized into static vs.~temporal phenotyping. Different ways to represent temporal data in a two-way array have been proposed in the past. Thus, when discussing temporal phenotyping, the structure will follow different data representation approaches to handle temporal information in a two-way context.

\subsection{Static Phenotyping}

Well-established matrix decomposition techniques, like PCA \citep{burgel2010clinical,georgiades2007structure,aliberti2016clinical,vavougios2016phenotypes} or NMF \citep{joshi2016identifiable}, have been a standard tool for phenotyping. Phenotypes and patient-clustering for a plethora of different diseases and conditions have been studied in the past, ranging from chronic obstructive pulmonary disease (COPD) \citep{burgel2010clinical, rennard2015identification} to autism disorder \citep{georgiades2007structure} as well as sleep apnoe \citep{vavougios2016phenotypes}. Papers that fit into this categorization, typically begin with a \emph{patients} $\times$ \emph{clinical attributes} matrix (see Figure \ref{fig:matrix-decomp}) and proceed to cluster patients using the factor matrices \citep{burgel2010clinical, vavougios2016phenotypes, aliberti2016clinical}. In \citep{wang2019enhancing}, phenotypes related to chronic lymphocytic leukemia are extracted. Although longitudinal patient data is used in this study, the resulting phenotypes are not temporal, as the longitudinal information is collapsed by summing along the temporal axis. The resulting matrix is a \emph{patient} $\times$ \emph{clinical features} matrix that contains count data and is decomposed using NMF. 

Various advances to PCA and NMF-based approaches have been introduced in the context of static phenotyping. For instance, in order to align phenotypes using NMF with target comorbidities, a \emph{supervision constraint} that enforces phenotype definitions to have non-zero entries according to a known list of comorbidities has been introduced \citep{joshi2016identifiable}. Besides, generalized low-rank models (GLRM) have been used to model mixed data, i.e., tabular data where each column follows a different statistical data type \citep{schuler2016discovering}.

\subsection{Temporal Phenotyping}

Given clinical features for a cohort over time, matrix decompositions cannot be applied readily as such temporal data cannot be represented as a two-way array. In the temporal phenotyping literature, this problem has been tackled using different approaches which can be categorized under: \textit{concatenation} \citep{hassaine2020learning, schuler2016discovering} and \textit{augmentation} \citep{luo2016predicting,ding2021unsupervised,stroup2019phenotyping}.

The most straightforward way to employ matrix factorizations for temporal phenotype extraction is to concatenate matrices along an axis. For instance, in \citep{hassaine2020learning}, \emph{age} $\times$ \emph{conditions} matrices for all patients are concatenated along the time/age axis to form one `skinny' matrix for the whole cohort. A different concatenation approach is used in \citep{schuler2016discovering}, where for each patient a vector is formed recording the counts of medical concepts in 6-month intervals, i.e., the same set of medical features are recorded over time. The resulting matrices are then approximated by low-rank methods, i.e., NMF in \citep{hassaine2020learning}, and Poisson PCA \citep{collins2001generalization}, which is tailored to count data using a loss similar to the KL-divergence based loss in (\ref{eq:KL}), in \citep{schuler2016discovering}.

The second approach to handle temporal data via matrix decompositions is sub-graph augmented non-negative matrix factorization (SANMF). Temporal information is encoded into a weighted graph. SANMF was introduced by \citep{luo2016predicting} with the specific use-case of analyzing physiologic time series data, and then also employed for unsupervised phenotyping of sepsis \citep{ding2021unsupervised} and multiple organ dysfunction \citep{ stroup2019phenotyping}. In \citep{luo2016predicting}, the time series of a physiological measurement is converted into a discretized representation that indicates whether the measurement is within the reference range or one or two standard deviations above or below it. These discretized values can be interpreted as weights in a \emph{time series graph}. The adjacent nodes in this time series graph are connected by three different kinds of edges, representing up, down, or no change. For each physiological measurement a corpus of time series graphs is formed that contains all different trends for all patients. Applying this to all physiological measurements leads to corpora of time series graphs. Frequent subgraph mining \citep{nijssen2005gaston} is then used to find subgraphs that repeatedly occur for patients. The subgraphs are encoded in a \emph{patients} $\times$ \emph{subgraphs} matrix. The first $n$ columns in this matrix could, for example, encode frequent subgraphs found for the first physiological variable. The resulting matrix is approximated using NMF to reveal patient groups and subgraph (trends) groups.

\section{Tensor Factorizations for Phenotyping}
\label{sec:tensorfactorizations}

\begin{figure*}[t]
\centering
\includegraphics[width=0.95\textwidth]{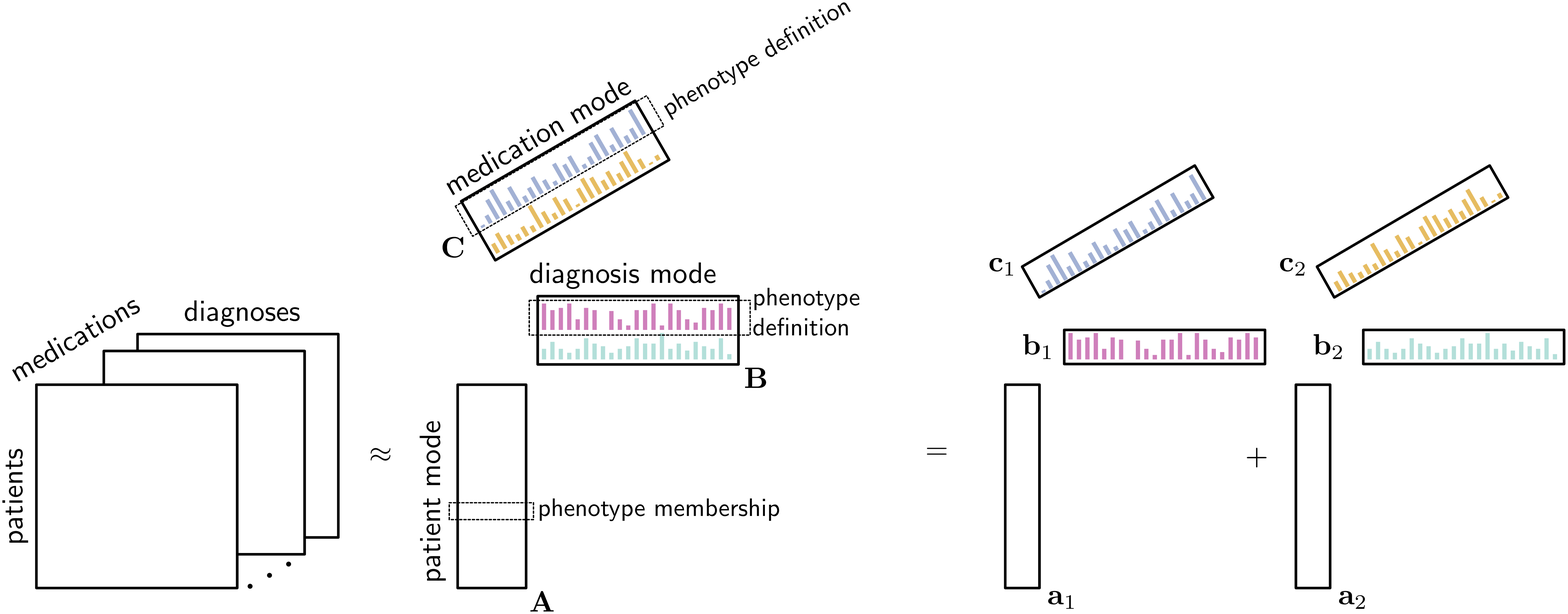}
\caption{Static phenotyping. A common \emph{patients} $\times$ \emph{medications} $\times $ \emph{diagnoses} tensor for static phenotyping. Each entry corresponds to the medication prescribed for a certain diagnosis and patient. In other words, this structure captures the co-occurrence of patients, diagnoses and medications. A \textit{phenotype} is defined as the combination of a corresponding medication-mode and diagnosis-mode component.}
\label{fig:static-tensor}
\end{figure*}

Data that has a multi-way structure, e.g., the co-occurrence of patients, diagnoses and medications, can be naturally represented with a (data) tensor. The advantage of this approach is that multi-modal interactions can be modeled and revealed. Most studies in this category shape the data into regular-shaped tensors, where each slab has the same dimensions \citep{wang2015rubik, ho2014marble, kim2017discriminative, perros2018sustain}. However, different-sized slabs, i.e., the dimension of one mode (typically time) varies across slabs, are increasingly studied in recent years to incorporate the temporal aspect. Temporal irregularity arises, for instance, when features for subjects are recorded during multiple clinical visits, which may greatly vary across the cohort \citep{perros2017spartan, yin2020logpar, afshar2020taste, ren2020robust, afshar2018copa}. For regular-shaped tensors, the CP model is mostly used, while for irregular shaped tensors, which are often used for temporal phenotyping, the PARAFAC2 model is common (see Table \ref{tab:literature}).

We categorize studies using tensor factorizations for EHR-based phenotyping under static phenotyping vs.~temporal phenotyping. 
On the highest categorization level within the subsections of static and temporal phenotyping, we first differentiate between CP based vs.~non-CP based models and PARAFAC2 based vs.~non-PARAFAC2 based models, respectively. We then follow a general structure that is based on different regularizations/constraints as this reflects the main advancements and contributions.

\subsection{Static Phenotyping} \label{sec:static_phenotyping}
For static phenotyping, the data is often arranged as third-order tensors recording diagnosis-prescription co-occurrences for a cohort as in Figure \ref{fig:static-tensor}  \citep{ho2014marble, wang2015rubik, perros2018sustain, kim2017discriminative, ho2014limestone}. While in most cases co-occurrence \textit{counts} are used \citep{henderson2018phenotyping, henderson2017granite, he2019distributed, ho2014marble}, in some cases, tensors with binary values are constructed \citep{wang2015rubik}. Note that while most studies integrate count data over a certain time window, they are considered under \textit{static phenotyping} because temporal information is not present in the resulting phenotypes \citep{ho2014limestone, ho2014marble,kim2017discriminative, wang2015rubik}. 

The constructed tensors are then often analyzed using a CP model, which has shown promising results in terms of uncovering clinically relevant phenotypes \citep{ho2014marble, wang2015rubik}. As tensors with co-occurrence counts are the most common way to represent static multi-way EHR data, KL-divergence is used when fitting the CP models in many instances (see Table \ref{tab:literature}).

In the first studies that analyzed such \emph{patient-diagnosis-medication} tensors, non-negative tensor factorizations based on a CP model were proposed, and coined \textsc{Limestone} and \textsc{Marble} \citep{ho2014limestone,ho2014marble}, respectively. Compared to \textsc{Limestone}, \textsc{Marble} introduces an offset tensor, and approximates the observed data using both an offset tensor and a so-called signal tensor. The offset part accounts for the common baseline characteristics in the population while the signal tensor models the phenotypes. Non-negativity has been used in many follow-up studies \citep{wang2015rubik, henderson2017granite, henderson2018phenotyping} as it facilitates interpretability. Within this subsection, we categorize the papers on the basis of their constraints that enforce sparsity, distinctive phenotypes, incorporate prior knowledge or label information. 

Sparsity of phenotypes is a desirable property as it makes interpretation easier by pushing less important values in the factor matrix towards zero. Generally, a typical way to enforce sparsity is to introduce an $\ell_1$ penalty term. Note, however, that introducing sparsity when using KL-divergence is challenging because the `inadmissible zeros’ \citep{chi2012tensors}, that can result from sparse factor matrices, make the objective ill-defined (due to the logarithm). In order to avoid this issue, an observed tensor can be factorized into a rank-one \emph{bias} (or offset) tensor with strictly positive entries and a \emph{signal tensor} that captures the phenotypes. Using this approach, the argument of the logarithm is always strictly positive. This way of modelling count data tensors is used, for instance, in \textsc{Marble} \citep{ho2014marble} and \textsc{Granite} \citep{henderson2017granite}. In \textsc{Marble}, sparsity in the phenotype factor matrices is enforced using simplex constraints together with a threshold parameter constraining the feasible space for the values in the factor matrices to be either zero or above a certain threshold
while in \textsc{Granite} an $\ell_2$ penalty term together with a simplex constraint is used.

\begin{figure*}[h!]
\centering
\includegraphics[width=0.95\textwidth]{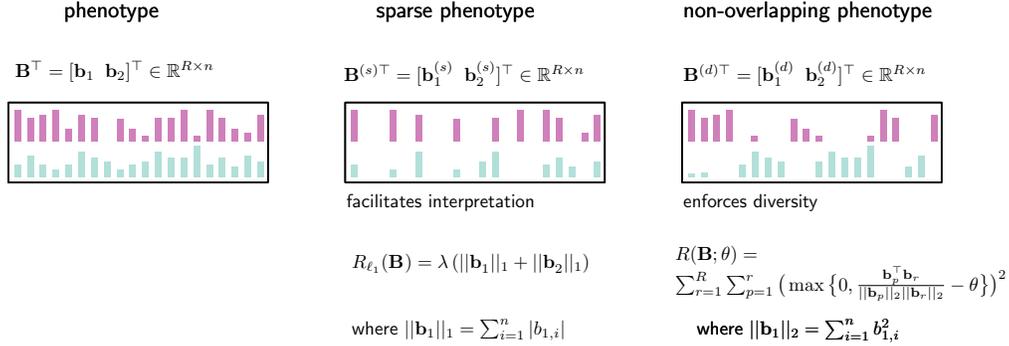}
\caption{Schematic phenotypes with different regularizations. The interpretability of phenotypes can be improved by enforcing sparsity $\M{B}^{(s)}$ or diversity $\M{B}^{(d)}$. For instance, in order to enforce a more diverse (i.e., non-overlapping) phenotype, an angular regularization term can be added when fitting a CP model, e.g., $ \min_{\M{A}, \M{B}, \M{C}} \norm{\mathcal{X} -\KOp{\M{A},\M{B},\M{C}}}_{F}^2 + \lambda (R(\mathbf B; \theta) + R(\mathbf C; \theta))$, where $\mathcal{X}$ is a \emph{patients}  - \emph{diagnoses}  - \emph{medications} tensor, $\M{A},\M{B}, \M{C}$ correspond to the factor matrices in \emph{patients}, \emph{diagnoses} and \emph{medications} modes; $\lambda$ is the penalty parameter, and $\theta$ defines the threshold above which the angle between two components is penalized.} 
\label{fig:phenotypes}
\end{figure*}

More distinct phenotypes with less overlap is another way to make interpretation easier and allow for a more targeted care. Less overlap can be enforced by penalizing non-orthogonality as in \textsc{Rubik} \citep{wang2015rubik} or by a more flexible angular regularization term (see Figure \ref{fig:phenotypes}) that is enforced on the factor matrices as in \citep{henderson2018phenotyping} or \textsc{(S)Granite} \citep{henderson2017granite, he2019distributed}. With the same goal of making phenotypes less overlapping, \citep{kim2017discriminative} introduce a similarity matrix that clusters similar phenotypes together. The similarity matrix is computed by Word2Vec, a neural network based method for word embeddings, using diagnoses and prescription sequences.

With unsupervised approaches, it is usually desirable that subgroups emerge without label information. Some studies, however, leverage a-priori medical knowledge \citep{wang2015rubik, henderson2018phenotyping} or label information \citep{kim2017discriminative, yang2017tagited} to enforce separability of different classes of patients. In \citep{henderson2018phenotyping}, prior knowledge is integrated by using a \textit{cannot-link} matrix so that patients with different disease status, e.g., cases vs. controls, are observed in different phenotypes. \textsc{Rubik} \citep{wang2015rubik} also incorporates prior medical knowledge via \textit{guidance constraints} such that factor matrices revealing phenotypes are constrained to be similar to a set of known features for certain diseases.
\citep{kim2017discriminative} propose to adapt the objective function to include a logistic regression term that enforces discriminative power by including label information, such that the phenotypes are separated according to mortality. In a similar way, \citep{yang2017tagited} propose a \textit{predictive task guided} tensor decomposition model by incorporating label information and a discriminative model as a penalty term.

A further distinct line of research is more concerned with the privacy-preserving computation of phenotypes \citep{kim2017federated, ma2019privacy}. The scenario under consideration is to compute phenotypes coming from multiple hospitals without sharing patient data. \textit{Federated} tensor factorization is proposed by \citep{kim2017federated} where patient-mode matrices are updated locally and feature-mode matrices are shared on a server, where a \textit{harmonized} global feature-mode matrix is computed and then sent back. The research carried out by \citep{ma2019privacy} has the same privacy-preserving goal, but differs mainly in three points: (i) by the usage of a more communication-efficient algorithm to solve the global consensus problem between the server and local sites, (ii) by the flexibility to allow hospital-specific factors via $\ell_{2,1}$ norm, i.e., by inducing sparsity in the patient-mode matrix, (iii) by giving differential privacy guarantees, i.e., a formal and rigorous guarantee that individual-level patient data cannot be deduced from patient factor matrices.

\paragraph{Non-CP based models for static phenotyping}

All previously mentioned papers in this subsection use CP (together with different constraints) as a model to discover static phenotypes. There are two exceptions \citep{yin2020learning, luo2015subgraph}. In \citep{luo2015subgraph}, subgraph augmented non-negative tensor factorization (SANTF) is introduced for discovering lymphoma subtypes from clinical texts by using a Tucker decomposition \citep{Tu66}. Each entry in the tensor is the co-occurrence count of a patient, a subgraph that encodes a common relation between medical concepts and a word. A Tucker model, more precisely referred to as Tucker3, is more flexible than the CP model, and approximates a third-order tensor $\T{X} \in \Real^{I \times J \times K}$ as follows, using a $(P,Q,R)$-component model:
\begin{equation*}
   \mathcal X \approx \sum_{p=1}^P\sum_{q=1}^Q\sum_{r=1}^{R}  g_{pqr} \mathbf a_p \, \circ \, \mathbf b_q \, \circ \, \mathbf c_r =   \KOp{\T{G};\M{A},\M{B},\M{C}},
\end{equation*}
where $\T{G} \in \Real^{P \times Q \times R}$
is the core tensor, $\M{A} \in \Real^{I \times P}, \M{B} \in \Real^{J \times Q}, \M{C} \in \Real^{K \times R}$ are the factor matrices corresponding to each mode. Tucker decompositions are not unique without further constraints. \citep{luo2015subgraph} use non-negativity constraints in all modes as well as for the core tensor. Uniqueness properties of such non-negative Tucker decompositions are yet to be understood \citep{ZhCi15}. In \citep{yin2020learning}, the co-occurrence assumption, which states that e.g.~diagnoses and medications that occur at the same clinical examination correspond to each other, is critically examined. It is stated that while this assumption holds for certain data sets, it might not hold for all. Thus, the \textit{true interaction-tensor} is assumed to be unknown, and only the marginal matrices (e.g., a \emph{patients} by \emph{diagnoses} matrix, a \emph{patients} by \emph{medications} matrix) are given. Factor matrices are estimated by maximizing the likelihood of marginalizations of the underlying unknown interaction-tensor. This model was coined Hidden Interaction Tensor Factorization (HITF) and is distinct from a CP model. Experimental evidence indicates that HITF-based approaches produce meaningful phenotypes that are sparser and more diverse compared to \textsc{Marble}, \textsc{Rubik}, and \textsc{Granite}.

\subsection{Temporal Phenotyping}

In recent years, temporal phenotyping using tensor decompositions has been studied. The first work that proposed to use PARAFAC2 to handle temporal irregularity was \textsc{Spartan} \citep{perros2017spartan}, to the best of our knowledge. Since then, it has been the most popular method to model temporal irregular EHR tensors. Applying PARAFAC2 to large-scale data sets has been a challenge due to computational issues \citep{bro1997parafac}. Scalable PARAFAC2 \textsc{Spartan} \citep{perros2017spartan} was developed to render PARAFAC2 feasible for large-scale problems by exploiting the sparsity structure. 
This efficient approach has also been used in \textsc{Copa} (Constrained PARAFAC2) \citep{afshar2018copa} where the PARAFAC2 model is fitted with additional constraints. Moreover, efforts have been made to make the temporal phenotype discovery more robust \citep{ren2020robust}, and more interpretable by enforcing a variety of different constraints such as sparsity \citep{afshar2018copa}, non-negativity \citep{afshar2018copa, perros2019temporal} and temporal smoothness \citep{afshar2018copa, yin2020logpar}. These advancements will be discussed in more detail in the following.

Non-negativity and sparsity are standard constraints that, as described in Section \ref{sec:static_phenotyping} for CP models, improve phenotype interpretability. It is, therefore, not surprising that one or both of these constraints are also incorporated when fitting PARAFAC2 models \citep{afshar2018copa, yin2020logpar}.

Better interpretability of temporal phenotypes is achieved by constraints that enforce temporal smoothness constraints \citep{yin2020logpar,afshar2018copa}. Temporal smoothness of the time-mode is a desirable property as it means both less fitting to noise and improved interpretability. The interpretation of smooth trajectories is easier because it is assumed that clinical parameters or disease severity do not change abruptly. Smoothness in time was enabled by enforcing the temporal factors to be linear combinations of M-spline basis functions \citep{afshar2018copa}, which form a non-negative basis that is defined piecewise by polynomials, inspired by earlier work carried out by \citep{helwig2017estimating, timmerman2002three}, or by penalizing the difference between the entries for consecutive visits in the time-mode factors \citep{yin2020logpar}.

For robustness to missing and erroneous entries, the \textsc{Repair} framework \citep{ren2020robust} proposes robust PARAFAC2 by modelling the tensor using a low-rank tensor and a sparse error tensor similar to the way robustness is introduced in matrix factorizations \citep{CaLi11}. The low-rank part relies on the PARAFAC2 model together with a new low-rank regularization function through nuclear norm constraints on the factor matrices. 
It has been demonstrated that \textsc{Repair} outperforms both \textsc{Spartan} \citep{perros2017spartan} and \textsc{Copa} \citep{afshar2018copa} in terms of model fit in the presence of missing and erroneous entries.

\paragraph{Non-PARAFAC2 based models for temporal phenotyping}
While PARAFAC2 is the most widely used method to handle temporal irregularity in EHR, another line of research proposes what could be called patient-level decomposition \citep{yin2019learning, zhou2014micro}. Collective non-negative tensor factorization (CNTF) \citep{yin2019learning} corresponding to a coupled CP model, and the \emph{shared basis approach} through coupled matrix factorization (CMF) in \citep{zhou2014micro} model the temporal dynamics of each patient separately while the phenotype definition is shared across the cohort. 

Given two matrices $\M{X} \in \Real^{I \times J_1}$ and $\M{Y} \in \Real^{I \times J_2}$ coupled in the first mode, e.g., \emph{clinical features} by \emph{visits} matrices are recorded for two patients for the same set of clinical features, CMF can be formulated as \citep{SiGo08a}:
\begin{equation} \label{eq:CMF}
   \min_{\mathbf A, \mathbf B, \mathbf C } \norm{\smash{\M{X} - \M{A}\M{B}\Tra}}_F^2 + \norm{\smash{\M{Y} - \M{A}\M{C}\Tra}}_F^2 
\end{equation}

where $\M{A} \in \Real^{I \times R}$ corresponds to the factor matrix in the clinical features mode. This formulation can be extended to jointly analyze a higher-order tensor and a matrix or multiple tensors, referred to as coupled matrix and tensor factorizations (CMTF) \citep{acar2011all} or coupled tensor factorizations. In \citep{yin2019learning}, for instance, \emph{time-labtest-medication} tensors for a cohort of patients
are jointly analyzed using a coupled CP model, where tensors are coupled via the \emph{labtest} and \emph{medication} modes.

Temporal regularization is enforced via a recurrent neural network (RNN) for temporal dependency of consecutive disease states. This means that the RNN is part of the objective function, and is jointly learned (for each patient separately) together with the coupled CP model. 

A different form of temporal irregularity is considered in \citep{zhang2021feature}, where the varying sampling frequency between features, some of which are measured almost continuously (e.g., heart rate), while others only very sporadically (e.g., blood test results), is addressed. Dynamic time warping (DTW) is used for the computation of pairwise distances between all temporal features for the whole patient cohort. In this way, a regular \textit{pairwise distance tensor} where each slab encodes the pairwise DTW-distances between all patients for one feature, is formed and decomposed by a CP model.

In \citep{zhao2019detecting} temporal irregularity does not arise and cardiovascular disease (CVD) is modeled via a CP model of a disease-patient-time tensor. The tensor is shaped in such a way that it records the time 10 years before a CVD event for each patient.


\setlength{\tabcolsep}{4pt}
\begin{sidewaystable}\small
\newgeometry{left=0cm,right=0cm,bottom=0cm}

\caption{Papers considered for this survey. Abbreviations used within this table: \textbf{Validation approach columns:} SA - Survival analysis; H - hypothesis testing; SL - supervised learning; L - Literature comparison; M - Medical expertise; \textbf{T column is for statistical data type:} r - real data; c - count data; b - binary data; m - mixed data; \textbf{Modes column:} p - patient; \clinical - clinical features; d - diagnosis (codes); t - time; \med - medication; proc - procedure; sg - subgraph; r - regions; w - weeks; y - years. In the model column, subscripts such as $KL$ (Kullback-Leibler), $log$ (logistic) or $I$ (informational) denote loss functions. If no subscript is used, the standard squared Frobenius norm ($Frob$) is used within the objective function.}\label{tab:literature}
\begin{tabular}{@{}lllllllllll@{}}\toprule
Reference & Model & SA & H & SL & L & M & Target (Disease)/case study & T & Modes & Focus/main contribution  \\  \midrule
\multicolumn{8}{l}{--- \textbf{matrix decomposition: static phenotyping} ---}  \\
\row{\cite{burgel2010clinical}}{PCA}{}{}{}{}{\y}{COPD}{m}{\p $\times$ \diverse}{application}
\row{\cite{georgiades2007structure}}{PCA}{}{\y}{}{\y}{}{autism spectrum disorder}{m}{\p $\times$ \diverse}{application}
\row{\cite{aliberti2016clinical}}{PCA}{}{\y}{}{\y}{}{bronchiectasis}{m}{\p $\times$ \diverse }{application}
\row{\cite{vavougios2016phenotypes}}{PCA+opt. scaling}{}{\y}{}{}{}{sleep apnea}{m}{\p $\times$ \clinical}{application}
\row{\cite{rennard2015identification}}{FA}{\y}{}{}{}{\y}{COPD}{m}{\p  $\times$ \diverse }{application}
\row{\cite{joshi2016identifiable}}{NMF$_{I}$}{}{}{\y}{}{\y}{ICU mortality}{c}{\p $\times$ \di }{weak supervision}
\row{\cite{wang2019enhancing}}{NMF}{}{}{\y}{}{\y}{lymphocytic leukemia}{c}{\p $\times$ \diverse}{application}
\row{\cite{schuler2016discovering}}{GLRM}{}{}{}{}{}{hospitalization}{m}{\p $\times$ \diverse}{application}
\multicolumn{11}{l}{--- \textbf{matrix decomposition: temporal phenotyping} ---}  \\
\row{\cite{luo2016predicting}}{SANMF}{}{}{\y}{}{}{ICU mortality}{c}{\p $\times$ \sg}{SANMF}
\row{\cite{ding2021unsupervised}}{SANMF}{}{\y}{}{}{}{sepsis}{c}{\p $\times$ \sg} {application} 
\row{\cite{stroup2019phenotyping}}{SANMF}{}{}{\y}{}{}{multiple organ dysfunction}{c}{\p $\times$ \sg}{application}
\row{\cite{hassaine2020learning}}{NMF$_{KL}$}{}{}{}{\y}{}{general multi-morbidity}{r}{age $\times$ disease per p}{temporal concatenation} 

\row{\cite{schuler2016discovering}}{PCA$_{Poisson}$}{}{\y}{}{}{}{autism spectrum disorder}{c}{\p $\times$ t}{application} 
\midrule 
 \multicolumn{11}{l}{--- \textbf{tensor decomposition: static phenotyping} ---}  \\
 \row{\cite{ho2014marble}}{CP$_{KL}$}{}{}{\y}{}{}{high cost beneficiaries}{c}{\p $\times$ \di $\times$ \proc}{sparse non-negative CP}
\row{\cite{ho2014limestone}}{CP$_{KL}$}{}{}{\y}{}{\y}{heart failure}{c}{ \p $\times$ \di $\times$ \pres }{sparse non-negative CP}
\row{\cite{yang2017tagited}}{CP$_{KL}$}{}{}{}{}{\y}{hospitalization/expenses}{c}{\p $\times$ \di $\times$ \pres}{supervision}
\row{\cite{henderson2017granite}}{CP$_{KL}$}{}{}{\y}{}{\y}{hypertension}{c}{\p $\times$ \med $\times$ \di}{distinct phenotypes}
\row{\cite{he2019distributed}}{CP$_{KL}$}{}{}{\y}{\y}{}{general multi-morbidity}{c}{\p $\times$ \di $\times$ \med }{distributed computation}
\row{\cite{henderson2018phenotyping}}{CP$_{KL}$}{}{}{\y}{}{\y}{hypertension/diabetes}{c}{\p $\times$ \di $\times$ \pres }{semi-supervision}
\row{\cite{perros2018sustain}}{CP}{}{}{}{}{\y}{heart failure}{c}{\p $\times$ \di $\times $ \med}{integer-constrained factors}
\row{\cite{kim2017discriminative}}{CP}{}{}{\y}{}{}{ICU mortality}{c}{\p $\times$ \di $\times$ \pres }{supervision, distinct phenotypes}
\row{\cite{kim2017federated}}{CP}{}{}{}{}{\y}{general multi-morbidity}{c}{\p $\times$ \med $\times$  \di }{privacy-preserving}
\row{\cite{ma2019privacy}}{CP}{}{}{\y}{}{}{ICU mortality}{c}{\p $\times$ \di $\times$ \proc }{privacy-preserving}
\row{\cite{wang2015rubik}}{CP}{}{}{}{}{\y}{general multi-morbidity}{b}{\p $\times$ \di $\times$ \med}{guidance constraints}
\row{\cite{yin2020learning}}{HITF}{}{}{\y}{}{\y}{ICU mortality}{c/r}{\p $\times$ \di $\times$ \med $\times$ \lab}{hidden interaction tensor}
\row{\cite{luo2015subgraph}}{SANTF}{}{}{}{\y}{\y}{lymphoma}{c}{\p $\times$ \sg $\times$ words}{SANTF}
 \multicolumn{11}{l}{--- \textbf{tensor decomposition: temporal phenotyping} ---} \\
 \row{\cite{zhou2014micro}}{CMF}{}{}{\y}{}{}{congestive heart failure}{c}{feature $\times$ t}{densification of EHR}
 \row{\cite{he2019distributed}}{CP$_{KL}$}{}{}{}{\y}{}{flu patterns}{c}{r $\times$ w $\times$ y}{distributed computation}
 \row{\cite{zhao2019detecting}}{CP}{\y}{\y}{}{\y}{\y}{cardiovascular disease}{b}{\p $\times$ disease $\times$ t}{application}
 \row{\cite{zhang2021feature}}{CP}{}{}{\y}{}{}{sepsis/acute kidney injury}{r}{\p $\times$ \p $\times$ feature}{DTW-CP}
 \row{\cite{yin2020logpar}}{PARAFAC2$_{log}$}{}{}{\y}{}{}{ICU mortality/heart failure}{b}{\p $\times$ \clinical $\times$ t }{logistic PARAFAC2}
\row{\cite{perros2019temporal}}{PARAFAC2}{\y}{}{}{}{\y}{medically complex children}{b}{\p $\times$ \clinical $\times$ t }{application}
\row{\cite{perros2017spartan}}{PARAFAC2}{}{}{}{}{\y}{medically complex children}{c}{\p $\times$ (\di \med) $\times$ t}{scalable PARAFAC2}
\row{\cite{ren2020robust}}{PARAFAC2$_{Frob+ l_1}$}{}{}{\y}{}{\y}{ICU mortality}{c}{\p $\times$ \di $\times$ t}{robust PARAFAC2}
\row{\cite{afshar2018copa}}{PARAFAC2}{}{}{}{}{\y}{general multi-morbidity}{b?}{\p $\times$ \clinical $\times$ t }{temporal smoothness}
 \row{\cite{yin2019learning}}{CNTF$_{KL}$}{}{}{\y}{}{\y}{ICU mortality}{b}{t $\times$ \med $\times$ \lab per p}{CNTF \& RNN regularization}
 \multicolumn{11}{l}{--- \textbf{coupled matrix/tensor decomposition} ---}  \\
\row{\cite{afshar2020taste}}{PARAFAC2/MF}{}{}{\y}{}{\y}{heart failure}{b?}{\p $\times$ (\di \med) $\times$ t}{temporal \& static information}
\row{}{}{}{}{}{}{}{}{m}{\p $\times$ \clinical}{}
\bottomrule
\end{tabular}
\end{sidewaystable}

\restoregeometry


\section{Validating Phenotypes}
\label{sec:validation}

The approaches for phenotype validation, i.e., the assessment of their clinical significance, can be grouped into \textit{internal} and \textit{external} categories. External validation means to bring in the expertise from a medical expert/practitioner or to collate the found phenotypes with the medical literature. Internal validation means to stay within data and assess the phenotypes using statistical testing (e.g., between cases and controls) or by using the extracted phenotypes in a subsequent supervised learning task. 

Internal and external validation can complement each other. In conjunction, they can establish strong evidence for clinical relevance \citep{joshi2016identifiable, wang2019enhancing, zhao2019detecting, henderson2018phenotyping, ho2014limestone, yin2019learning, yin2020learning}, e.g., when both a predictive task and medical expertise independently agree that a certain phenotype is important.

Many papers considered in this survey use both internal and external validation. Table \ref{tab:literature} shows the validation approach for each paper. 
We observe that validation methods based on a prediction task and medical expertise are the most common ones. Moreover, the combination of a prediction task and the consultation of medical expertise is also quite common. In the following, we discuss internal and external validation approaches in more detail.

\subsection{Internal Validation}

\paragraph{Survival Analysis}

In survival analysis \citep{bewick2004statistics}, the \textit{time until an event occurs} is studied. Typically, as the term \textit{survival} indicates, this event is death, but it can also be the onset of a disease. The term \textit{time-to-hazard} is used to refer to this notion. Compared to supervised learning, survival analysis is a less common way for phenotype validation \citep{perros2019temporal, rennard2015identification, zhao2019detecting}. After the decomposition, the patient-mode can be understood as grouping of patients, for which \textit{survival functions} can be computed. The survival function is a non-increasing function that shows the cumulative survival times for the cohort. In this way, differences between phenotypes can be analyzed. In \citep{zhao2019detecting}, for example, where cardiovascular disease is studied, \textit{Kaplan-Meier} models are presented that show the survival functions of six different patient subgroups, taking myocardial infarction as the event. Kaplan-Meier plots are a descriptive tool that can indicate differences between subgroups. Using a log-rank test, statistical significance between different survival functions can be assessed. Thus, survival functions can directly link different subgroups of patients to survival statistics and thereby assess the clinical relevance of the uncovered phenotypes.

\paragraph{Hypothesis Testing.} Given that the research questions and available data allow it, hypothesis tests can help to identify differences between subgroups \citep{ding2021unsupervised,schuler2016discovering, zhao2019detecting}. Hypothesis tests can, for instance, be performed in case-control studies, where the different uncovered phenotypes (score matrix) are compared between cases and controls. 
In \citep{ding2021unsupervised}, where the objective is to uncover sepsis phenotypes, two sample $t$-tests are used to assess differences between the three different phenotypes/subgroups. Demographic information and clinical features of the subgroups are compared against each other; for instance, age significantly differs between the subgroups, as well as the clinical feature \textit{fluid electrolyte imbalance}. Performing statistical tests between all subgroups and features can show fine-grained differences between phenotypes.

\paragraph{Supervised Learning.}

A predictive task subsequent to the extraction of phenotypes makes quantitative evaluations and comparisons possible. The \textsc{Pacifier} framework \citep{zhou2014micro}, for instance, decomposes incomplete features $\times$ time matrices to estimate missing entries and uncover phenotypes. They call this process \textit{densification}. In order to validate the phenotypes and  methodology, the completed matrix is used to predict heart failure via a sparse logistic regression, and compared with other baselines (e.g.,~interpolation). Applying a logistic regression to the phenotype membership matrix has been used for validating phenotypes for hypertension \citep{henderson2018phenotyping, henderson2017granite}, high-cost beneficiaries \citep{ho2014marble}, mortality prediction \citep{luo2016predicting, yin2019learning, yin2020learning, ma2019privacy} and heart failure \citep{ho2014limestone, afshar2020taste}. Supervised learning as a subsequent step to an unsupervised decomposition is a common way that can indicate clinical plausibility. However, it is important to note that a downstream predictive task does not measure interpretability which is why medical experts are often consulted in addition.

\subsection{External Validation}

In cases where neither a control group (or any other subdivision of the cohort) nor labeled data is present, the only option is external validation which will be discussed in the following.

\paragraph{Comparison with existing literature}

Existing research, even when performed using different cohorts or study designs, might supply evidence for the clinical relevance of phenotypes. Relating a phenotype to what is already known in the medical literature is used in some studies using matrix and tensor decomposition methods for phenotyping \citep{zhao2019detecting, hassaine2020learning, he2019distributed}. In \citep{zhao2019detecting}, for instance, where the disease under consideration is the cardiovascular disease, known comorbidities from the scientific literature are linked with the uncovered phenotypes.

\paragraph{Medical expertise}

Medical expertise is crucial for methodological and clinical validation. Using uncovered phenotypes for some supervised learning task, as described above, only quantifies the predictive power with respect to some target. However, it is a different question whether the phenotypes are clinically meaningful and easy to interpret. Including clinical experts to evaluate the phenotypes under these criteria is a common practice (see Table \ref{tab:literature}). Typically, clinicians are asked to assess the phenotypes on a certain scale, e.g.~1) clinically meaningful, 2) possibly clinically meaningful, and 3) not clinically meaningful \citep{henderson2018phenotyping, wang2015rubik, ho2014limestone, henderson2017granite, yin2020learning} or 1) poor, 2) fair, 3) good, 4) excellent \citep{joshi2016identifiable}.

A further common practice is to let medical experts label the uncovered phenotypes \citep{wang2019enhancing,perros2017spartan, ho2014limestone, afshar2018copa}. This means that a medical expert examines the latent concepts (phenotype definitions) and finds succinct descriptions.

In one case, medical expertise was used for model selection \citep{wang2019enhancing}, i.e., NMF is run using different number of components, and then evaluated with respect to the clinical relevance of the uncovered latent concepts.

\begin{table*}[h]
\caption{Commonly used datasets: Sutter \citep{choi2017using}, CMS, MIMIC-II \citep{saeed2011multiparameter}, MIMIC-III \citep{johnson2016mimic}  } 
\begin{tabular}{@{}lll@{}}\toprule
Dataset & Data set description & Used in \\
 \midrule
Sutter  & Palo Alto Medical Foundation Clinics; & \citep{yin2020logpar} \\
& medication and diagnosis information   & \citep{afshar2020taste} \\
& from 50 to 80 year old adults in a  heart & \citep{perros2018sustain} \\
& failure study & \\
CMS & 3 years of claim records synthesized & \citep{yin2020logpar, ren2020robust},  \\ 
& from 5\% of the 2008 Medicare population & \citep{perros2018sustain, afshar2020taste} \\
MIMIC-II  & Physiologic data and vital signs time series & \\ & collected from tens of thousands & \citep{luo2016predicting} \\
&   of ICU patient monitors   & \\
MIMIC-III  & successor of MIMIC-II $^1$  & \citep{joshi2016identifiable, ren2020robust} \\
& & \citep{kim2017discriminative} \\
& & \citep{ding2021unsupervised} \\
& & \citep{yin2020learning} \\
& & \citep{kim2017federated} \\
& & \citep{ma2019privacy} \\
& & \citep{yin2019learning} \\
& & \citep{he2019distributed} \\
& & \citep{yin2020logpar}\\
& & \citep{zhang2021feature}\\
\bottomrule
\end{tabular}\label{tab:datasets} \\
\footnotesize{$^1$ see https://mimic.mit.edu/docs/ for the documentation of MIMIC-II and MIMIC-III as well as for differences between the two databases.}
\end{table*}

\section{Discussion}

Static and temporal phenotyping via low-rank approximation-based methods have been studied to gain a better understanding of raw EHR. The major advantages of these methods are their model transparency and the ability to uncover (temporal) phenotypes that are explainable and interpretable, as well as their potential to reveal novel phenotypes. In this section, we discuss some limitations that frequently arise within the phenotyping literature and lay out possible future directions.

\subsection{Challenges and Limitations}
There are common challenges and limitations with matrix and tensor factorization-based approaches concerning model uniqueness, computational reproducibility, number of components, and the validation of components (or phenotypes). 

\subsubsection{Uniqueness}
A general advantage of tensor factorizations such as CP and PARAFAC2 over matrix factorizations is their uniqueness guarantees (up to permutation and scaling ambiguities). A model that has no unique solution is arbitrary. The computed factor matrices (or phenotypes) could be meaningless as they change in different runs even when the same solution, i.e., the same cost function value, is obtained. Thus, uniqueness is an essential requirement for a factorization model if the goal is to interpret the extracted components. Matrix factorizations such as NMF are not unique in general \citep{laurberg2008theorems}. However, enforcing additional sparsity can potentially make the factorization unique. Therefore, in order to provide certainty that the interpreted model and the extracted phenotypes are not arbitrary, it is crucial to discuss and investigate the uniqueness properties of matrix and tensor factorization-based approaches as in several phenotyping studies \citep{perros2019temporal, yin2020logpar, perros2017spartan, afshar2020taste}.

\subsubsection{Computational reproducibility}
While essential uniqueness is crucial for the interpretability of factors, that does not alone guarantee computational reproducibility. Here, with computational reproducibility, we refer to obtaining the same factors (e.g., phenotypes) using the same data and the same method \citep{AdKaAk22}. When fitting matrix/tensor factorization-based models, we often solve a non-convex optimization problem. Therefore, due to local minima, components from even essentially unique models, e.g., CP and PARAFAC2, can be spurious when the methods are initialized using different initializations.
To minimize the chance of ending up with spurious components from a local minimum, i.e., different phenotypes in different runs, the optimization procedure needs to be run multiple times using different (random) initializations and the factors corresponding to the best run, e.g., the one with the lowest cost function value \citep{perros2019temporal}, the most consistent run, should be interpreted \citep{AdKaAk22}. Despite being important for interpretability, computational reproducibility, best run selection and model stability have only been addressed in a few studies in matrix/tensor factorization-based computational phenotyping \citep{ho2014limestone, perros2019temporal, AdKaAk22}.

\subsubsection{Number of components}
An appropriate low-rank model separates the systematic part of the data from the residuals that arise due to model and measurement errors. The number of components is commonly chosen based on whether an additional component significantly improves the model fit. However, determining the number of components is a challenging task, and various methods to determine the model order, i.e., the number of components, have been studied such as imputation error or cross-validation-based approaches \citep{louwerse1999cross, udell2016generalized}, core consistency diagnostic for CP and PARAFAC2 models \citep{BrKi03, perros2019temporal} as well as Bayesian approaches relying on automatic relevance determination \citep{MoHa09}.
Choosing the appropriate number of components is discussed in some phenotyping studies \citep{zhao2019detecting, perros2019temporal, perros2018sustain}, and the sensitivity of the performance to the number of components has been discussed \citep{yin2020logpar, afshar2020taste}. Model order selection needs to be carried out carefully; otherwise, uncovered patterns might be spurious and non-reproducible. While not commonly used, choosing the number of components based on model stability, i.e., by using the cross-correlation of factors from different runs \citep{wu2016stability}, and computational reproducibility can be a sensible and supplementary approach next to other common selection criteria. Within EHR-based phenotyping, model stability is used to the determine the number of components in \textsc{Sustain} \citep{perros2018sustain}.

\subsubsection{Validation of components}
It is crucial to assess the clinical relevance of a phenotype. Survival analysis has been successfully employed for internal validation. However, it has been used only in a few studies \citep{perros2019temporal, rennard2015identification, zhao2019detecting}. 
In contrast to supervised learning, survival analysis has the potential to uncover fine-grained information about the progression of a disease, or different subgroups defined by a phenotype. Kaplan-Meier plots, although being purely descriptive, can indicate the severity of a phenotype definition, and thus, be of great therapeutic value.

\subsection{Possible Future Directions}
There are several future directions that have the potential to advance computational phenotyping further.

\begin{figure*}[ht]
\centering
\includegraphics[width=0.7\textwidth]{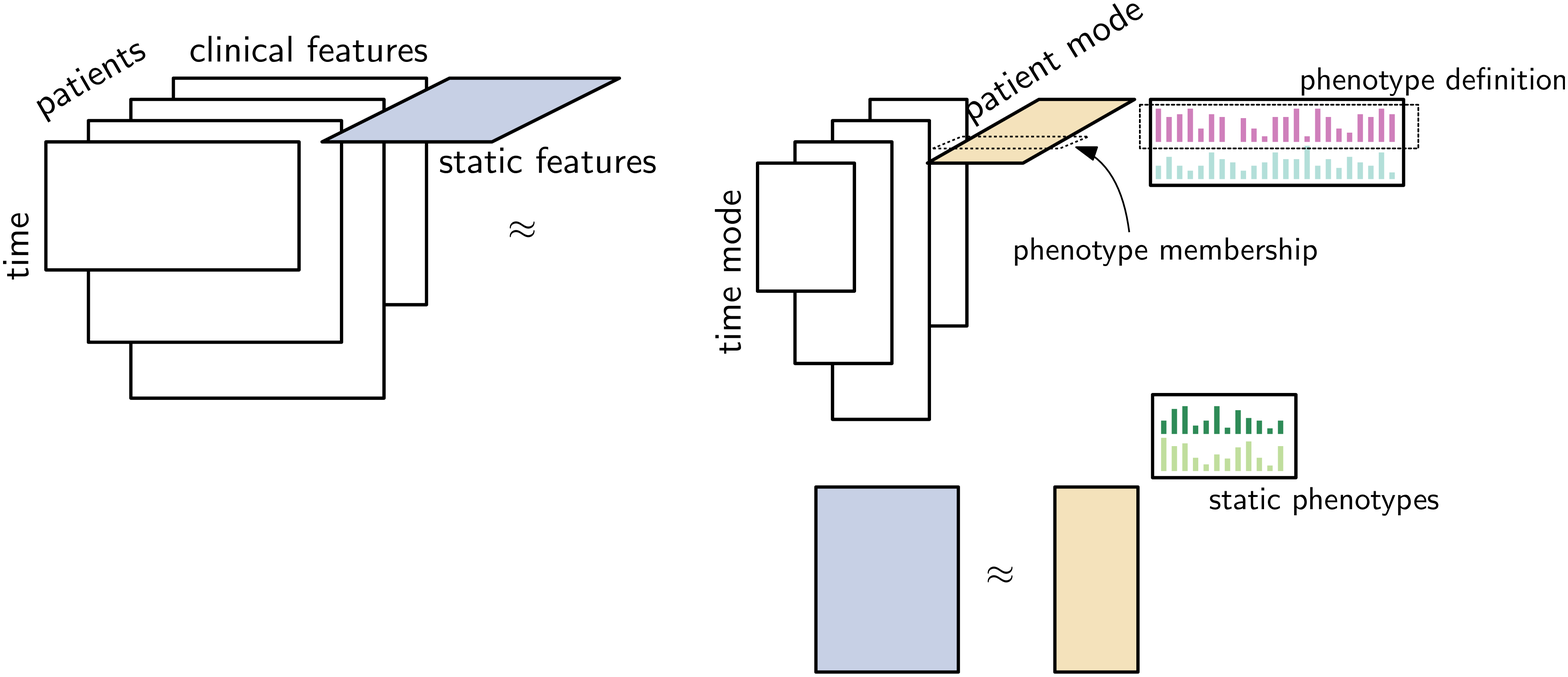}
\caption{Coupled matrix and tensor factorization (CMTF). Static and temporal information is, for instance, coupled via the patient mode. This approach reveals static and temporal phenotypes while it can potentially also uncover more meaningful patient groupings.} \label{fig:CMTF}
\end{figure*}

\sloppy{First, jointly analyzing data from different modalities has the potential to reveal better phenotypes, and has shown promise in precision medicine \citep{PrMaEa17}. While different data modalities can, in some cases, be represented as a single data tensor \citep{LuAh17}, often there is a need to jointly analyze data sets in the form of multiple matrices and higher-order tensors, for instance, to fuse temporal as well as static data sources \citep{afshar2020taste,GuThPa21} (as illustrated in Figure \ref{fig:CMTF}), to jointly analyze multiple modalities \citep{LiNaLu20} or for joint analysis of data from controls and patients \citep{YiCh21}. Coupled matrix and tensor factorizations (CMTF) have been effective tools for joint analysis of such data sets in data mining \citep{acar2011all, AcBrSm15, PaFaSi16}, in particular, in multi-modal neuroimaging data analysis \citep{AcShLe19,ChKoPa22} as well as recommender systems \citep{ZhCaZhXiYa10,ErAcCe15}. In EHR-based phenotyping, CMTF-based approaches may allow for the incorporation of both temporal and non-temporal sources. The patient mode, for instance, can be enforced to be a low-rank representation of the temporal as well as the static information. Fusing these different data sources can lead to an improved clustering of patients while uncovering both temporal and static phenotype definitions. So far, CMTF-based phenotyping has been limited focusing only on EHR data \citep{afshar2020taste}, where temporal diagnosis and medication data and static features have been jointly analyzed. Future work may extend EHR-based phenotyping by incorporating other modalities such as various omics data sets, and jointly analyze those data sets using CMTF-based approaches. Recent advances in CMTF methods (e.g., different loss functions for different data sets, various constraints \citep{schenker2020flexible}) may facilitate the progress in that direction.}

Second, handling of different data types need further attention. Different statistical data types can be either ignored \citep{luo2016predicting, afshar2018copa} or modeled via adapting the loss functions \citep{yin2020logpar, henderson2018phenotyping}. However, a question that does not seem to be conclusively answered is how the choice of loss functions influences the discovery of phenotypes. While there is some indication that `appropriate' loss functions outperform `non-appropriate' ones \citep{schuler2016discovering}, there is also evidence that applying the standard Frobenius norm to e.g.~count data results in meaningful phenotypes \citep{luo2016predicting, becker2022phenotyping}. For a more extreme case, consider a third-order tensor with mixed statistical data types across the feature-mode. That is, the tensor records, for instance, different laboratory parameters together with the disease severity over time. This means that there is one tensor structure that has multiple different statistical data types. For the matrix case, this has already been studied \citep{udell2016generalized, schuler2016discovering}. However, it has not yet been extended to the tensor case.

Finally, privacy-preserving phenotyping is still an under-researched field. While being studied for CP-models \citep{ma2019privacy}, it has not yet been applied to (irregular) temporal EHR data.
 
\section{Conclusion}

In this paper, we provided a comprehensive review of low-rank approximation-based approaches for computational phenotyping and outlined how common challenges such as irregularity along the time mode or overlapping phenotypes are tackled. The appeal of low-rank methods lies in their well-understood theoretical foundations as well as in their model transparency. Our survey shows that CP and PARAFAC2 are common low-rank models for static and temporal phenotyping, respectively. Their uniqueness guarantees make them a suitable tool as the factors are non-arbitrary. Different constraints such as sparsity and non-negativity or regularizations such as an angular penalty to enforce more distinct phenotypes have been used with the ultimate goal to make phenotypes more interpretable and useful in clinical contexts. The validation of phenotypes is a major challenge that has been addressed in different ways. The most common being a subsequent supervised learning task or validation by medical expertise. We outlined that the literature would benefit from including a discussion about model uniqueness and computational reproducibility. Finally, we discussed coupled matrix and tensor factorizations, and the handling of different statistical data types as a possible future directions.

\section*{Funding Information}
This work is part of the \textit{DeCipher} project that is funded by the Research Council of Norway. 

\section*{Conflict of Interest}
The authors declare that they have no conflicts of interest.

\bibliography{temporal-phenotyping.bib}

\end{document}